\documentclass[pdflatex,sn-mathphys-num]{sn-jnl}

\usepackage{graphicx}%
\usepackage{multirow}%
\usepackage{amsmath,amssymb,amsfonts}%
\usepackage{amsthm}%
\usepackage{mathrsfs}%
\usepackage[title]{appendix}%
\usepackage{xcolor}%
\usepackage{textcomp}%
\usepackage{manyfoot}%
\usepackage{booktabs}%
\usepackage{algorithm}%
\usepackage{algorithmicx}%
\usepackage{algpseudocode}%
\usepackage{listings}%

\theoremstyle{thmstyleone}%
\theoremstyle{thmstyletwo}%

\theoremstyle{thmstylethree}%

\begin{document}

\title[Article Title]{Single Image Rain Streak Removal Using Harris Corner Loss and R-CBAM Network}


\author[1]{\fnm{Jongwook} \sur{Si}}\email{jwsi425@kumoh.ac.kr}

\author*[2]{\fnm{Sungyoung} \sur{Kim}}\email{sykim@kumoh.ac.kr}

\affil[1]{\orgdiv{Dept. of Computer·AI Convergence Engineering}, \orgname{Kumoh National Institute of Technology}, \orgaddress{\street{Daehak-ro 61}, \city{Gumi}, \postcode{39177}, \state{Gyeongbuk}, \country{Republic of Korea}}}

\affil*[2]{\orgdiv{School of Computer Engineering}, \orgname{Kumoh National Institute of Technology}, \orgaddress{\street{Daehak-ro 61}, \city{Gumi}, \postcode{39177}, \state{Gyeongbuk}, \country{Republic of Korea}}}


\abstract{The problem of single-image rain streak removal goes beyond simple noise suppression, requiring the simultaneous preservation of fine structural details and overall visual quality. In this study, we propose a novel image restoration network that effectively constrains the restoration process by introducing a Corner Loss, which prevents the loss of object boundaries and detailed texture information during restoration. Furthermore, we propose a Residual Convolutional Block Attention Module (R-CBAM) Block into the encoder and decoder to dynamically adjust the importance of features in both spatial and channel dimensions, enabling the network to focus more effectively on regions heavily affected by rain streaks. Quantitative evaluations conducted on the Rain100L and Rain100H datasets demonstrate that the proposed method significantly outperforms previous approaches, achieving a PSNR of 33.29 dB on Rain100L and 26.16 dB on Rain100H. In particular, additional ablation studies confirmed that the introduction of Harris Corner Loss plays a critical role in enhancing restoration quality, ensuring both structural consistency and fine detail preservation even in complex rain streak scenarios. This research presents a new approach that comprehensively addresses both noise removal and structural preservation, offering high scalability and practicality as a fundamental technology applicable to a wide range of future image restoration and enhancement tasks.}




\maketitle

\section{Introduction}\label{sec1}

\indent
Artificial intelligence and computer vision technologies have become essential components in a wide range of engineering applications, including autonomous vehicles, unmanned aerial vehicles (UAVs), surveillance systems, smart cities, medical imaging, remote sensing, and human-robot interaction~\cite{bib1, bib2}. These systems rely heavily on visual data to accurately perceive and make decisions in real time about their external environments. Consequently, the quality of input images significantly affects the overall performance and reliability of such systems. However, real-world environments are rarely ideal, and adverse weather conditions such as rain, fog, and snow are among the major factors that degrade image quality.

Among these, images captured during rainy conditions often contain rain streaks, which appear as semi-transparent, linear patterns superimposed on the image~\cite{bib3, bib4}. These streaks blur the contours of the background and objects, distort texture and structural information, and ultimately degrade the accuracy of computer vision algorithms. For instance, autonomous vehicles must accurately detect lanes, pedestrians, and traffic signs; yet, the presence of rain streaks can lead to false or missed detections, potentially resulting in fatal accidents. Likewise, UAVs may fail to identify terrain features or obstacles during aerial navigation, and surveillance systems may mistake rain streaks for moving objects or fail to detect actual intruders. Thus, the rain streak problem extends beyond simple visual distortion—it poses a significant challenge to the safety and performance of real-world engineering systems. As a result, removing rain streaks from images has emerged as a critical technical task~\cite{bib5, bib6, bib7}.

Rain streak removal is a highly challenging problem due to the diverse and dynamic characteristics of rain patterns~\cite{bib8, bib9}. Rain streaks can vary in direction, length, transparency, and density from frame to frame, and they are often difficult to distinguish when the visual contrast with the background is low. Recently, deep learning-based approaches have been actively explored to tackle this problem~\cite{bib10, bib11, bib12}. Architectures such as UNet, ResNet, GANs, and Transformers have shown impressive results in image restoration tasks and have been applied to rain streak removal. Nevertheless, existing methods still face several limitations.

First, many networks are designed with a strong focus on removing rain streaks, often at the cost of preserving structural information. As a result, the restored images tend to appear blurred or lack clear object boundaries. Second, typical CNN architectures process all regions of an image uniformly, which limits their ability to focus on visually important regions or object-centric areas. Third, most loss functions are based on pixel-wise differences (e.g., L1 or L2), which fail to capture human-perceived visual quality and structural consistency. Fourth, due to high model complexity and large computational requirements, many existing models are unsuitable for real-time applications or deployment on resource-constrained platforms such as embedded systems.

To address these limitations, we propose \textbf{SHARK}, denoting \textbf{S}ingle image rain streak removal using \textbf{HA}rris corner loss and \textbf{R}-cbam networ\textbf{K}, as a novel rain streak removal network that integrates the R-CBAM Block with a loss function based on Harris corner responses. The proposed model is built upon the UNet architecture and incorporates R-CBAM Blocks to improve both learning stability and feature enhancement. The residual structure facilitates the effective flow of information in deep networks, while CBAM (Convolutional Block Attention Module)~\cite{bib13} allows the model to focus more precisely on semantically significant areas within the image. This enables the network to effectively attend to the background and object boundaries, where rain streaks typically cause the most distortion, thereby enabling more accurate removal.

In addition, this study introduces a Harris Corner Loss to enhance the network’s ability to preserve structural information. Harris corner detection is a classical yet effective method for extracting structural features such as edges, contours, and textures in images. By incorporating Harris responses into the loss function, the network is guided to maintain structural consistency during the learning process—something that conventional pixel-wise losses tend to overlook. This approach prevents the degradation of edges and shapes when removing rain streaks and improves the perceptual quality of the restored images.

The proposed network combines the multi-scale processing capabilities of the UNet architecture, the attention-guided feature refinement provided by the R-CBAM Block, and the structure-preserving supervision enabled by Harris Corner Loss. This synergy enhances both the accuracy of rain streak removal and the perceptual clarity of the output. The main contributions of this study can be summarized as follows:

\begin{itemize}
    \item We employ R-CBAM Blocks to effectively extract critical features from complex rain-degraded images while enhancing learning stability through residual connections.
    \item We propose the Harris Corner Loss, which complements pixel-wise losses by enforcing the preservation of structural consistency and key edge features.
    \item The overall framework is designed to process and reconstruct fine-grained visual details through hierarchical feature integration, enabling effective reconstruction and generation of rain-free images.
\end{itemize}

\section{Related Works}\label{sec2}

Fu et al.~\cite{bib14} proposed a deep learning architecture based on residual learning to remove rain streaks from a single image. Their method separates the input image into low-frequency and high-frequency components and focuses on removing rain streaks from the high-frequency part. The proposed network emphasizes lightweight and efficient processing by optimizing specifically for rain streak removal without involving complex operations. The commonality with the proposed method is the utilization of residual learning to separate and eliminate rain streaks. However, while the previous work only aimed at removing rain streaks, the present study further incorporates Harris Corner-based structural preservation, thereby ensuring that fine structural details in the restored image are also maintained. Consequently, the prior method is advantageous for fast processing and low computational cost but has limitations in accurately restoring complex edges and detailed structures.

Li et al.~\cite{bib15} introduced a Context Aggregation Network that combines recurrent structures with Squeeze-and-Excitation (SE) blocks. Their approach progressively removes rain streaks through recursive iterations and enhances feature representations by adaptively recalibrating channel-wise responses. The similarity with the proposed method lies in the shared focus on enhancing feature representations to improve restoration quality. However, their method relies solely on channel importance without explicitly enforcing structural information preservation. As a result, while it successfully leverages channel importance for performance improvement, it remains relatively weak in preserving fine object contours and edges.

Fu et al.~\cite{bib16} proposed a lightweight network that extracts and merges features across multiple scales using a pyramid structure. This method emphasizes minimizing computational complexity and memory usage while effectively removing rain streaks by exploiting multi-scale information. The proposed method shares the idea of leveraging multi-scale feature extraction for effective rain removal. However, unlike the present study, it does not introduce explicit mechanisms for structural preservation, such as protecting high-frequency or contour information. Therefore, while this method achieves efficiency and lightweight design, it is relatively less effective in restoring intricate structural details.

Jiang et al.~\cite{bib17} proposed a network that progressively fuses features from different scales. Their design particularly emphasizes restoring high-resolution details to improve the visual quality after rain removal. The commonality with the proposed method is the emphasis on utilizing multi-scale features and enhancing fine detail restoration. However, while the proposed method further considers structural preservation based on Harris Corner Detection to ensure fine restoration of edges and textures, this previous work mainly focuses on feature fusion across scales. As a result, while it demonstrates strong restoration performance, it may still fall short in achieving the precision of edge restoration compared to the present approach.

Fu et al.~\cite{bib18} introduced a Deep Detail Network (DDN) that first restores a rain-free low-resolution image and then supplements high-frequency details. They focused on the observation that rain streaks mainly affect fine details and made high-frequency restoration the core of their network design. The similarity with the proposed method lies in the focus on high-frequency component restoration. However, unlike the proposed method that explicitly enforces structural preservation using a Harris Corner-based loss, this study merely aimed at enhancing fine details without introducing strong structural constraints. Therefore, while DDN improves rain removal at fine scales, it lacks the precision control over structure preservation present in the proposed approach.

Finally, Fan et al.~\cite{bib19} proposed a Residual-Guide Network (RGN) that explicitly models and predicts residual maps to guide the rain removal process. Their network relies on modeling residual information to enhance rain streak removal performance. The proposed method shares the use of residual information for rain streak removal. However, the proposed work not only removes residual rain streaks but also enforces structural consistency between the input and restored images using Harris Corner constraints, leading to more robust restoration performance. While RGN effectively removes rain regions, it does not fully address the preservation of detailed structures and boundaries, which the proposed method successfully improves.

In summary, these six previous studies have explored different directions for designing networks aimed at rain streak removal, primarily leveraging strategies such as residual learning, multi-scale feature processing, and high-frequency detail restoration. However, most of the prior works did not explicitly incorporate constraints for structural consistency preservation, especially regarding the restoration of object contours and high-frequency components. In contrast, the present study distinguishes itself by introducing quantitative constraints based on Harris Corner Detection to simultaneously enhance rain removal performance and visual quality, thereby offering a structurally faithful and perceptually superior deraining solution.

\section{Proposed Method}\label{sec3}

\subsection{Model Architecture}\label{subsec1}
\begin{figure}[h]
\centering
\includegraphics[width=0.9\textwidth]{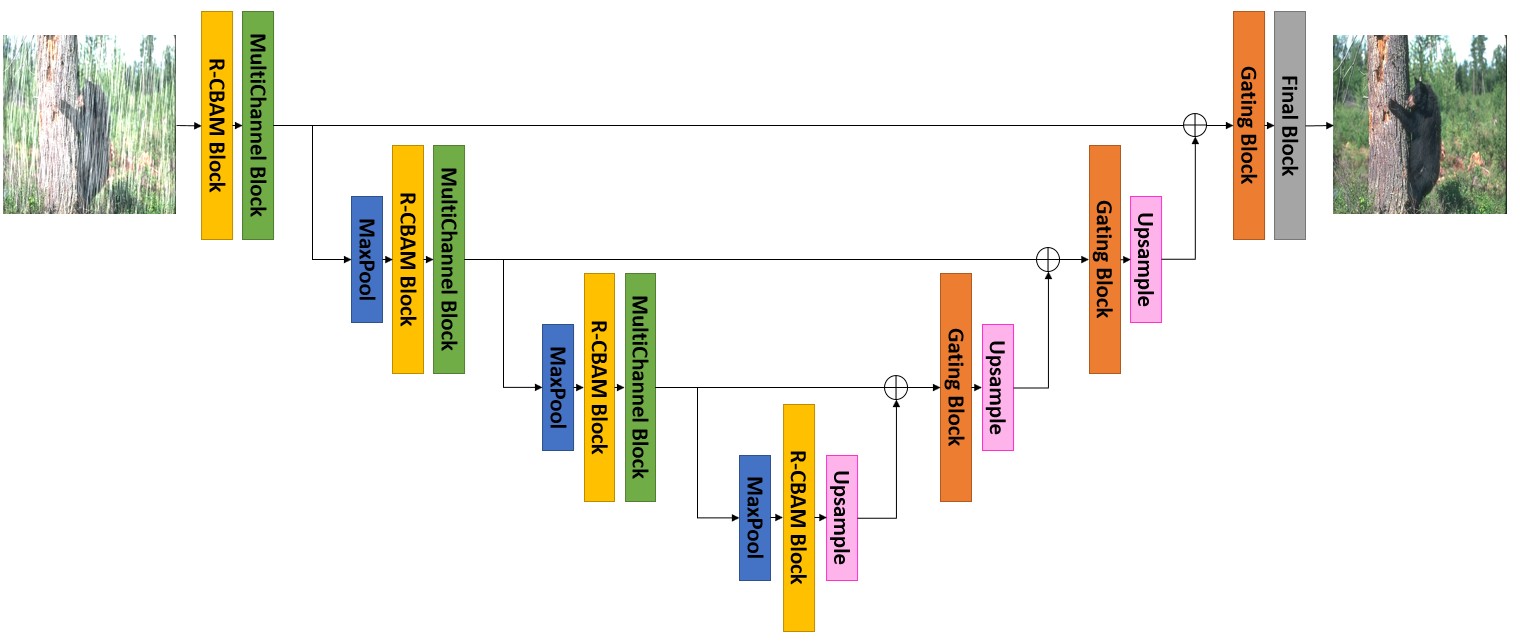}
\caption{Overall Architecture of the Proposed Deep Learning Network}\label{fig1}
\end{figure}

\indent
In this study, we propose a novel image restoration network to address the problem of rain streak removal from a single input image. The proposed model is based on the U-Net architecture and inherits its multi-scale feature aggregation and skip-connection-based information preservation capabilities. However, to enhance the removal of complex rain streak patterns and to preserve structural information without degradation, several architectural improvements are integrated into the baseline design. These improvements are organized into five major functional modules, each strategically embedded within the network architecture based on its computational role—namely feature abstraction, channel-wise refinement, structural reconstruction, spatial emphasis, and output normalization. The proposed architecture is illustrated in Figure~\ref{fig1}.

\begin{figure}[h]
\centering
\includegraphics[width=0.9\textwidth]{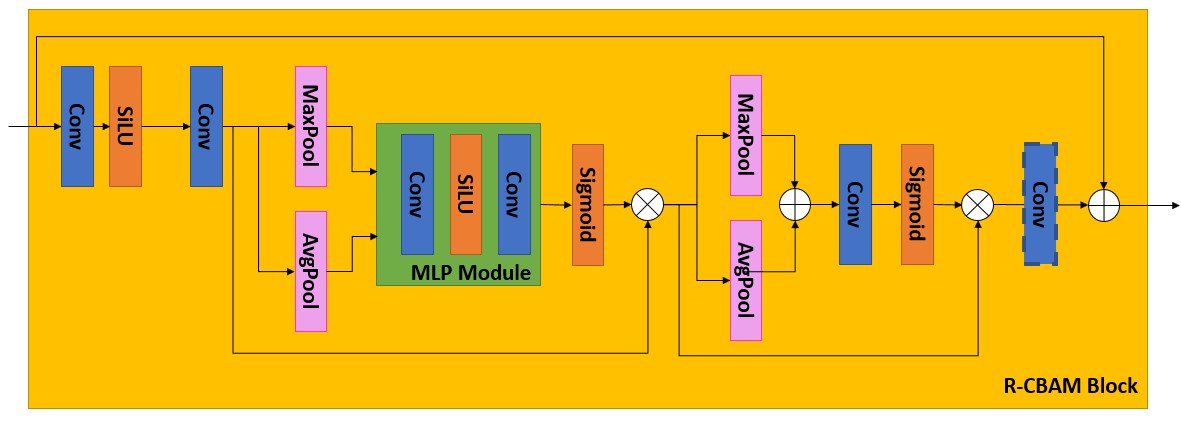}
\caption{Detailed Structure of the R-CBAM Block}\label{fig2}
\end{figure}

The first module performs hierarchical feature abstraction from the input image. Given an RGB image $I\in\mathbb{R}^{3\times H\times W}$, the input is passed through four encoder blocks, which project the spatial domain into increasingly abstract feature spaces. Each encoder block is composed of a R-CBAM Block, which combines the residual learning structure with the Convolutional Block Attention Module (CBAM). It is illustrated in Figure~\ref{fig2}. This design not only extracts features but also incorporates attention mechanisms to focus on regions frequently affected by rain streaks. The R-CBAM Block consists of two consecutive $3\times3$ convolutional layers followed by the SiLU (Sigmoid-weighted Linear Unit) activation function, and sequentially applies channel attention and spatial attention. The block processes the input feature map $X$ through two consecutive $3\times3$ convolutions and activations to generate an intermediate feature representation $F_1$ and its subsequent transformation $F_2$, as defined in Equation~\eqref{eq1}. 

\begin{equation}
F_1 = \text{SiLU}(\text{Conv}_{3\times3}(X)), \quad F_2 = \text{Conv}_{3\times3}(F_1)
\label{eq1}
\end{equation}

\indent
The channel-wise importance modulation module regulates the information flow by computing the relative contribution of each channel in the input feature map $F_2$. To achieve this, global average pooling and global max pooling are applied to $F_2$, generating two channel-wise statistical descriptors that discard spatial information. These descriptors are independently passed through a shared multilayer perceptron (MLP) composed of a sequence of $1\times1$ convolution, ReLU activation, and another $1\times1$ convolution. The outputs are summed and activated via a sigmoid function to produce the importance weight vector $M_c$, as formulated in Equation~\ref{eq:2}:

\begin{equation}
M_c(F_2) = \sigma(\text{MLP}(\text{AvgPool}(F_2)) + \text{MLP}(\text{MaxPool}(F_2)))
\label{eq:2}
\end{equation}

Here, $\sigma$ denotes the sigmoid activation. The resulting vector $M_c$ is element-wise multiplied with the original feature map $F_2$ to yield the attention-modulated output $F_c$, thereby enhancing informative channels and suppressing less relevant ones, as expressed in Equation~\ref{eq:3}:

\begin{equation}
F_c = M_c(F_2) \cdot F_2
\label{eq:3}
\end{equation}

The spatial significance mapping operates by computing both average and maximum projections of $F_c$ along the channel axis. These two spatial maps are concatenated and processed with a $7\times7$ convolution to generate a spatial modulation map $M_s$, as shown in Equation~\ref{eq:4}:

\begin{equation}
M_s(F_c) = \sigma(\text{Conv}_{7\times7}([\text{Avg}(F_c); \text{Max}(F_c)]))
\label{eq:4}
\end{equation}

The output $M_s$ is element-wise multiplied with $F_c$ to produce the spatially refined representation $F_{\text{cbam}}$, as defined in Equation~\ref{eq:5}:

\begin{equation}
F_{\text{cbam}} = M_s(F_c) \cdot F_c
\label{eq:5}
\end{equation}

Finally, a residual connection is applied by adding the input $X$ to $F_{\text{cbam}}$. If the number of channels in $X$ and $F_{\text{cbam}}$ are identical, direct addition is performed. Otherwise, a $1\times1$ convolution is used to match the dimensionality, as specified in Equation~\ref{eq:6}:

\begin{equation}
Y = F_{\text{cbam}} + 
\begin{cases}
X & \text{if } C_{\text{in}} = C_{\text{out}} \\
\text{Conv}_{1\times1}(X) & \text{otherwise}
\end{cases}
\label{eq:6}
\end{equation}

This statistics-based modulation technique selectively enhances meaningful features across both channel and spatial dimensions, while suppressing irrelevant information, thereby improving the representational capacity and output quality of the model.

\indent
These combined attentions enable the network to highlight boundaries and high-frequency regions associated with rain streaks and selectively learn meaningful representations even in early training stages. Moreover, the residual connections ensure smooth gradient flow and prevent vanishing gradient issues, facilitating deep abstraction without compromising the input information.

The second module, the MultiChannelBlock, is appended after each encoder output to enhance inter-channel representation diversity. This module comprises two sequential $3\times3$ convolutional layers with SiLU activations followed by a final $1\times1$ convolution. The design objective is to decompose and recognize complex rain streak patterns that vary in size, orientation, density, and transparency. While traditional convolution layers are limited in scope, the MultiChannelBlock strengthens inter-channel coupling and cascades different receptive fields to better isolate and represent high-frequency rain components. The $1\times1$ convolution effectively adjusts the channel dimensions and removes redundant features without increasing computational complexity. This module plays a critical role in maintaining robust performance across varying environments while ensuring network compactness and training efficiency.

The third module is dedicated to structural reconstruction and is implemented along the decoder path. Based on the high-dimensional features extracted by the encoders, this module restores the spatial resolution to match that of the input image. The decoder upscales each feature map via bilinear interpolation and concatenates it with the corresponding encoder feature map through skip-connections. The merged features are then processed by a R-CBAM Block, which further refines them. These skip-connections preserve spatial detail by bridging low-level and high-level features, minimizing structural distortion. By applying CBAM again within the decoder, the network reinforces attention on meaningful regions and suppresses rain-induced artifacts. Compared to traditional U-Net designs, this configuration achieves superior restoration fidelity, especially around object boundaries and streak-affected zones.

The fourth module enhances spatial discrimination using an GatingBlock applied after each decoder output. Although structurally simple, consisting of a single $1\times1$ convolution followed by a Sigmoid activation, which computes an gating map $\alpha$. This gating map is then multiplied with the decoder feature $F_{\text{dec}}$ to yield a refined output, as expressed in Equation~\eqref{eq7}.

\begin{equation}
F_{\text{refined}} = \alpha \cdot F_{\text{dec}}, \quad \alpha = \sigma(\text{Conv}_{1\times1}(F_{\text{dec}}))
\label{eq7}
\end{equation}

It computes an gating map representing the importance of each pixel position and multiplies it element-wise with the input feature map. This operation corrects potential omissions in fine structural information and eliminates residual rain streaks that may remain after decoding. Since the GatingBlock is lightweight, it contributes to the overall enhancement of visual quality without adding computational burden, ensuring that the final output is sharper and more visually natural.

Finally, the fifth module is the output normalization layer, which transforms the last decoder output into a three-channel RGB image. This is achieved using a $1\times1$ convolution followed by a Sigmoid activation function, which normalizes the output to the $[0,1]$ range. This normalization ensures numerical stability and allows the restored image to be interpreted as a valid visual representation. The Sigmoid activation also prevents excessive gradient propagation during early training, stabilizing the learning process and promoting faster convergence.

Each module in the proposed architecture is explicitly designed to fulfill a specific functional purpose and is embedded within the encoder-decoder structure to work synergistically. The R-CBAM Block extracts semantically rich features while enhancing focus through dual attention. The MultiChannelBlock introduces multi-dimensional representations to capture the variability of rain streaks. The decoder path, reinforced by skip-connections and CBAM, facilitates accurate structural recovery. The GatingBlock provides spatial correction, and the output normalization layer guarantees compatibility with standard visual formats. Collectively, these modules lead to a significant improvement in rain streak removal performance and ensure the generation of visually consistent and structurally accurate restored images. Furthermore, the architectural design supports high reliability, computational efficiency, and deployment scalability, making it a robust engineering solution for adverse weather vision restoration systems.

\subsection{Loss Functions}\label{subsec2}

In this study, a sophisticated loss function structure is designed to minimize the loss of structural information and maximize perceptual quality during the rain streak removal process. The proposed loss function is composed of three major components, each serving a complementary role: (1) L1 loss for reducing pixel-wise differences, (2) SSIM (Structural Similarity Index Measure) loss, which reflects perceptual similarity as perceived by humans, and (3) Harris Corner-based regularization loss for preserving structural consistency. The final loss function is formulated as a weighted summation of these three loss terms, guiding the network to effectively remove rain streaks while maintaining high-frequency and structural information.

The $L_1$ loss minimizes the absolute pixel-wise difference between the reconstructed image $\hat{I}$ and the ground-truth image $I$, and is defined as expressed in Equation~\eqref{eq:l1}.

\begin{equation}
L_1 = \| \hat{I} - I \|_1
\label{eq:l1}
\end{equation}

This loss term primarily contributes to the global brightness alignment and texture reconstruction, playing a key role in stabilizing the training process and accelerating convergence.

Pixel-wise similarity alone is insufficient to guarantee perceptual quality as recognized by human vision. To overcome this limitation, we incorporate a loss function based on the Structural Similarity Index Measure (SSIM), which jointly considers luminance, contrast, and structural components to assess visual similarity. The SSIM score ranges from 0 to 1, where higher values indicate greater similarity. For loss computation, we use the same equation as Equation~\eqref{eq:ssim}.

\begin{equation}
L_{\text{SSIM}} = 1 - \text{SSIM}(\hat{I}, I)
\label{eq:ssim}
\end{equation}

Rain streak removal goes beyond a simple denoising task; it critically depends on preserving the inherent visual information of the image, such as background structure, object contours, and textures. Rain streaks typically cause visual distortions in high-frequency regions of an image—namely, around object boundaries and edges that carry significant structural importance. Due to this characteristic, traditional pixel-wise loss functions such as $L_1$ or $L_2$ are limited in their ability to restore intricate and complex structural details effectively. To overcome this limitation, this study introduces a novel loss term based on the Harris Corner Detection algorithm, which is known to be highly responsive to structural features such as edges and corners in an image. Rather than merely guiding the model to remove rain streaks, this loss term acts as a crucial structural constraint that leads the network to preserve semantically important features. It ultimately plays a pivotal role in directing the overall learning behavior of the network.

The process of generating a corner map based on the Harris Corner method is as follows. For an input image, Sobel operations are applied to compute the gradients in the X and Y directions for each channel, resulting in gradient maps. Using these, the structure tensor is constructed at each spatial location and stabilized by applying a Gaussian smoothing operation $G$. Here, $G$ represents Gaussian smoothing using a $5\times5$ filter. Based on this tensor, the Harris corner response $R$ is calculated as expressed in Equation~\eqref{eq:harris_r}, where $\det(M) = I_{xx}I_{yy} - I_{xy}^2$, $\text{trace}(M) = I_{xx} + I_{yy}$. The response value $R$ reflects whether a pixel is located in a structurally significant region—larger values indicate areas near edges or corners.

\begin{equation}
R = \det(M) - k \cdot (\text{trace}(M))^2
\label{eq:harris_r}
\end{equation}

Next, a binary corner map $C$ is generated by thresholding the response value $R$ using a threshold $\tau$, as expressed in Equation~\eqref{eq:corner_map}.

\begin{equation}
C = \mathbf{1}(R > \tau \cdot \max(R))
\label{eq:corner_map}
\end{equation}

The resulting corner map $C$ represents the structurally important regions. It is computed for both the ground-truth image $I$ (yielding $C_{\text{input}}$) and the reconstructed image $\hat{I}$ (yielding $C_{\text{output}}$), and used to measure structural consistency between them. The Harris Corner loss is then defined as the $L_1$ distance between these two corner maps. It is expressed in Equation~\eqref{eq:harris_loss}. It ensures that the loss is concentrated only on structurally important regions, guiding the model to remove rain streaks without damaging edges and boundaries of objects. This approach significantly improves both training efficiency and structural fidelity. When structure is not preserved, structural quality metrics such as SSIM often drop drastically; this loss term effectively prevents such deterioration.

\begin{equation}
L_{\text{Harris}} = \| C_{\text{output}} - C_{\text{input}} \|_1
\label{eq:harris_loss}
\end{equation}

\begin{figure}[h]
\centering
\includegraphics[width=0.7\textwidth]{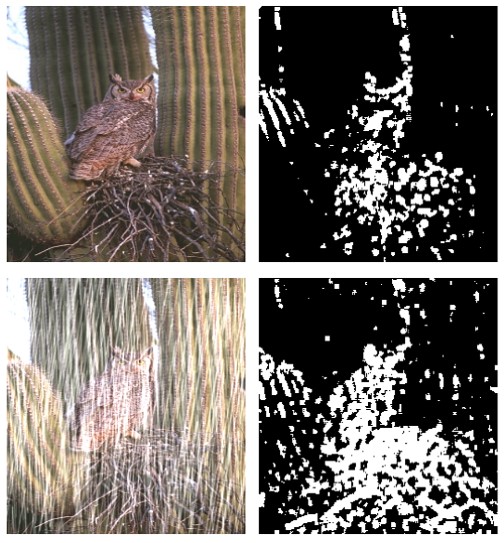}
\caption{Computation Results of the Harris Corner-Based Map (Input and Harris Corner Map)}\label{fig3}
\end{figure}

Figure~\ref{fig3} presents the Harris corner maps generated for images with and without rain streaks using the same parameter settings employed in this study. When rain streaks are prominent, they tend to be detected as corners, resulting in a high number of responses in the map. Therefore, assuming that the maps become similar once rain streaks are removed, this property can be utilized as a loss function.

Unlike $L_1$ or $L_2$ losses, which average pixel differences across the entire image and thus struggle with localized, directionally varying noise such as rain streaks, the Harris Corner-based loss provides several advantages. First, in terms of structural consistency, it selectively emphasizes visually critical features such as object contours and textures. Second, thanks to its localization sensitivity, it imposes the loss only at corner locations, enabling high-precision learning while reducing unnecessary computation. Third, it improves visual naturalness by preventing the image from becoming too blurred or smooth after the removal of the rain streak. Empirically, models trained with the Harris Corner loss demonstrate sharper object boundaries and better-preserved textures in the restored images. Furthermore, performance improvements are observed in structural quality metrics such as SSIM and PSNR. Since SSIM is directly related to structural consistency, this supports the effectiveness of the proposed loss term.

In summary, the Harris Corner loss serves as a core module that quantitatively reinforces the network's ability to preserve structure during rain streak removal. Not only is it vital for enhancing perceptual quality, but it also ensures engineering reliability in real-world image restoration applications.

To effectively train the proposed rain streak removal network, we design a composite loss function that enforces both pixel-level accuracy and structural consistency. Specifically, the loss function is composed of three main components: (1) a pixel-wise reconstruction loss, (2) a perceptual structural similarity loss, and (3) a structure-preserving regularization loss based on Harris corner response. The overall objective function is defined as Equation~\eqref{eq:total_loss}:

\begin{equation}
L_{\text{total}} = \lambda_1 L_1 + \lambda_2 L_{\text{SSIM}} + \lambda_3 L_{\text{Harris}}
\label{eq:total_loss}
\end{equation}

\section{Experiments}\label{sec4}
\subsection{Datasets}\label{subsec3}
\indent
The Rain100 dataset~\cite{bib20} is a widely adopted benchmark dataset designed to evaluate the performance of single-image rain removal algorithms. It consists of synthetically generated paired image samples that closely simulate real-world rainy conditions, where each rainy image is accompanied by its corresponding clean (ground-truth) image. This paired structure makes it highly suitable for supervised learning tasks and facilitates rigorous quantitative evaluation of deraining models.

Rain100 is divided into two subsets based on the intensity and complexity of rain streak patterns: Rain100L (Light Rain) and Rain100H (Heavy Rain). Each subset is further split into training and testing sets. Rain100L comprises relatively simple scenes with sparse and light rain streaks, while Rain100H includes more challenging scenes with dense, directional, and high-frequency rain patterns. This division allows researchers to assess the robustness of deraining algorithms under both low-complexity and high-complexity weather conditions. Specifically, the Rain100L sub-set contains 200 derain images and 200 rainy images for training, and 100 clean/rainy image pairs for testing. In contrast, the Rain100H subset is significantly larger and more challenging, offering 1800 Derain and 1800 rain images for training, and 200 derain and 200 rain images for testing.

In this study, both Rain100L and Rain100H are employed to comprehensively evaluate the proposed network under diverse rain conditions. By utilizing datasets with varying levels of complexity, we aim to verify not only the effectiveness of rain removal but also the model’s ability to preserve fine structural details such as object boundaries, edges, and textures. The Rain100 data set plays a crucial role in this research, as it enables a precise assessment of structural consistency in removing tasks and serves as a standardized benchmark to measure the general performance of the model and the visual fidelity in different scenarios.

\subsection{Environements}\label{subsec4}
\indent
In this study, the proposed Rain Streak removal model was trained and evaluated under a carefully constructed experimental environment. All experiments were conducted on a system running Ubuntu 18.04 LTS, equipped with two NVIDIA RTX 3090 GPUs. The implementation was carried out using Python 3.9 with the PyTorch deep learning framework, which provided a flexible and modular interface for designing the network architecture, defining loss functions, and orchestrating the training pipeline.

For optimization, the Adam optimizer was employed, with hyperparameters set to $\beta_1 = 0.9$ and $\beta_2 = 0.999$. The initial learning rate was configured as $1\times10^{-4}$.

To compute the Harris Corner loss, the Harris response coefficient was fixed at $k = 0.08$, and the threshold $\tau$ was set to $0.01$. These parameters were empirically chosen to enhance sensitivity to high-frequency regions, such as object edges and texture boundaries, enabling the loss term to focus effectively on structurally significant areas of the image.

Furthermore, the weighting coefficients for the overall loss function are determined empirically as follows: $\lambda_1 = 10$ for the $L_1$ loss, $\lambda_2 = 5$ for SSIM loss, and $\lambda_3 = 5$ for the Harris Corner loss.

All image data used during training were uniformly resized to a resolution of $256 \times 256$ pixels to ensure consistency in input dimensions and compatibility with the network architecture. The batch size was set to 4, optimizing GPU memory usage and training efficiency. The model was trained for a total of 500 epochs, and during each epoch, performance was monitored on a validation set using structural and perceptual quality metrics such as PSNR (Peak Signal-to-Noise Ratio) and SSIM (Structural Similarity Index Measure).

\subsection{Results Analysis}\label{subsec5}
\indent
In this paper, the performance of related works and the proposed method was evaluated on the Rain100L and Rain100H datasets. Figure 3 and 4 present qualitative comparisons demonstrating the restoration results of the proposed method on the Rain100L and Rain100H datasets, respectively. In each figure, the first column shows the rainy input images, the second column shows the restored images generated by the proposed method, and the third column shows the corresponding ground-truth (GT) images. the corresponding quantitative evaluation results are presented in Table 1.

\begin{figure}[h]
\centering
\includegraphics[width=0.9\textwidth]{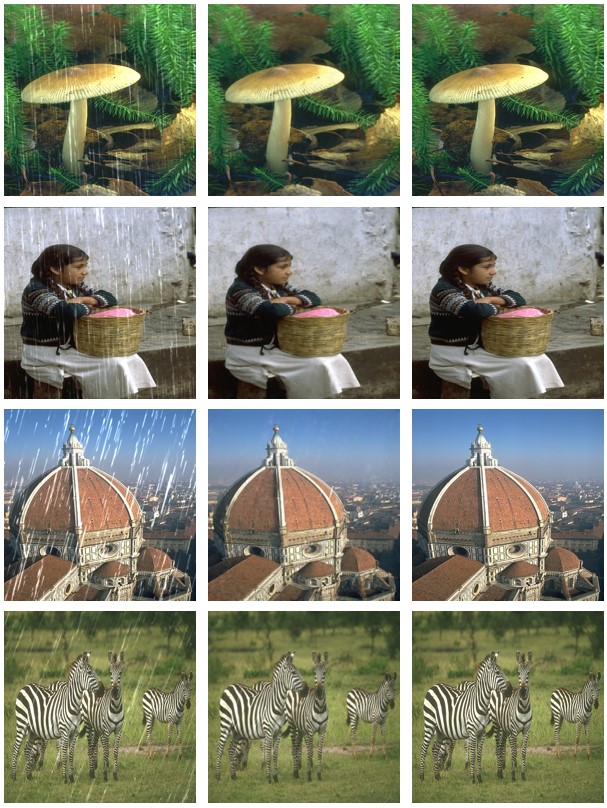}
\caption{Comparison of Results Using Rain100L Dataset (Input, Model's Output, Ground Truth)}\label{fig4}
\end{figure}

In Figure~\ref{fig4}, which presents the Rain100L results, the proposed method effectively removes various intensities of Rain Streaks while preserving the fine object structures and textures at a high level. For example, in the first row featuring a mushroom image, the edges and fine textures of the surrounding plants are sharply restored by the proposed method, closely matching the GT. In the second row with the child image, the facial features, clothing wrinkles, and the texture of the basket are naturally preserved without becoming blurred, clearly demonstrating that the Harris Corner-based structural loss effectively maintains critical object boundaries and detailed structures. Furthermore, in images such as the architectural dome and the group of zebras, Rain Streaks are successfully removed while fine structures and background textures are well preserved.

\begin{figure}[h]
\centering
\includegraphics[width=0.9\textwidth]{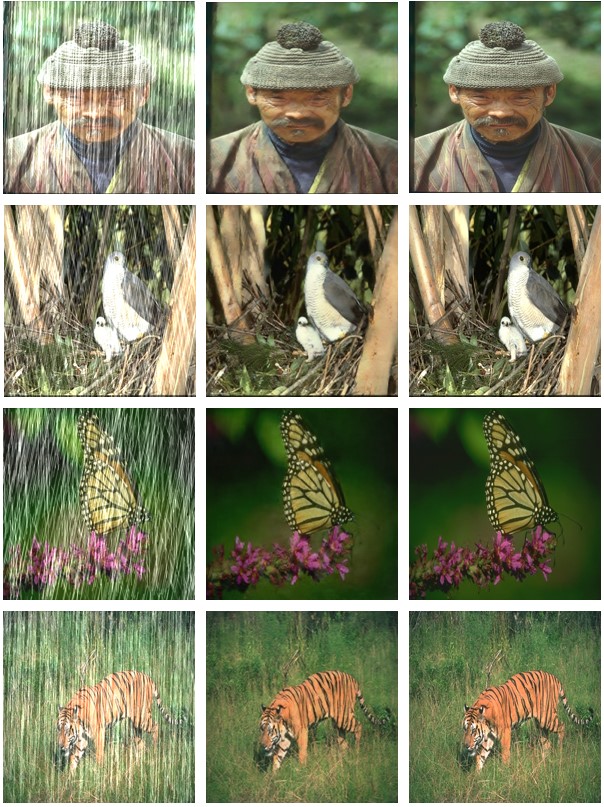}
\caption{Comparison of Results Using Rain100H Dataset (Input, Model's Output, Ground Truth)}\label{fig5}
\end{figure}

Figure~\ref{fig5} shows the results on the Rain100H dataset, which contains more complex and intense Rain Streaks. Although the input images are severely distorted by dense and heavy Rain Streaks, the proposed method effectively removes most of the rain artifacts while restoring the shape and fine details of the objects without degradation. For instance, in the first row with the portrait image, the texture of the hat, the wrinkles on the face, and the details of the clothing are naturally restored, appearing very similar to the GT. Similarly, in the second row with the bird image, the third row with the butterfly image, and the fourth row with the tiger image, the proposed method successfully removes rain streaks even when they are intricately intertwined with the background, while maintaining sharp object contours and fine structural details. In particular, there is almost no observable blurring at object boundaries or backgrounds, indicating that the proposed R-CBAM structure dynamically emphasizes important features from both spatial and channel perspectives, thereby significantly enhancing restoration performance.

Overall, Figures~\ref{fig4} and~\ref{fig5} visually demonstrate that the proposed Harris Corner-based structural loss and R-CBAM Block are highly effective not only in removing rain streaks but also in maintaining structural consistency and visual quality. In particular, the proposed method effectively overcomes common issues encountered in previous methods, such as loss of structural boundaries and degradation of texture details, generating more natural and sharp restoration results. This visually confirms the superiority of the proposed method, complementing the quantitative results presented previously.

In this study, the proposed method was quantitatively evaluated against prior works, including DerainNet~\cite{bib14}, RESCAN~\cite{bib15}, LPNet~\cite{bib16}, UMRL~\cite{bib17}, DDN~\cite{bib18}, and ResGuidNet~\cite{bib19}, using the Rain100L and Rain100H datasets. The detailed evaluation results are presented in Table~1. Based on the quantitative values, we conducted a comprehensive analysis while considering the methodological characteristics of each baseline.

\begin{table}[t]
\caption{Quantitative Comparison with Previous Works on Rain100L and Rain100H}
\centering
\begin{tabular}{lcccc}
\toprule
\textbf{Method} & \multicolumn{2}{c}{\textbf{Rain100L}} & \multicolumn{2}{c}{\textbf{Rain100H}} \\
\cmidrule(lr){2-3} \cmidrule(lr){4-5}
 & PSNR & SSIM & PSNR & SSIM \\
\midrule
DerainNet~\cite{bib14} & 25.39 & 0.874 & 14.85 & 0.443 \\
RESCAN~\cite{bib15} & 26.36 & 0.786 & 23.56 & 0.746 \\
LPNet~\cite{bib16} & 25.63 & 0.898 & 23.77 & 0.823 \\
UMRL~\cite{bib17} & 29.18 & 0.923 & 26.01 & 0.832 \\
DDN~\cite{bib18} & 32.16 & 0.936 & 21.92 & 0.764 \\
ResGuidNet~\cite{bib19} & 33.16 & \textbf{0.963} & 25.25 & 0.841 \\
\textbf{Ours} & \textbf{33.29} & 0.954 & \textbf{26.16} & \textbf{0.854} \\
\bottomrule
\end{tabular}
\end{table}

For the Rain100L dataset, the proposed method achieved a PSNR of 33.29 and an SSIM of 0.954, demonstrating the highest restoration performance overall. In comparison, DerainNet~\cite{bib14} recorded a PSNR of 25.39 and an SSIM of 0.874. Although DerainNet adopted residual learning to remove rain streaks, it did not introduce explicit constraints for preserving structural information, resulting in minor losses in high-frequency details. RESCAN~\cite{bib15} achieved a PSNR of 26.36 and an SSIM of 0.786; while it introduced a recurrent squeeze-and-excitation block to emphasize important features, its precision in restoring fine object boundaries remained limited. LPNet~\cite{bib16}, which focused on lightweight architecture through pyramid structures, reported a PSNR of 25.63 and an SSIM of 0.898; despite leveraging multi-scale information, it faced challenges in preserving detailed structures. UMRL~\cite{bib17} achieved a PSNR of 29.18 and an SSIM of 0.923, presenting relatively high performance but still lower than the proposed method due to the absence of explicit structural preservation mechanisms. DDN~\cite{bib18} recorded a PSNR of 29.18 and an SSIM of 0.923, emphasizing high-frequency detail restoration but showing residual rain streaks near object boundaries. ResGuidNet~\cite{bib19} achieved a PSNR of 33.16 and an SSIM of 0.963. Although ResGuidNet slightly surpassed the proposed method in SSIM by 0.009, the proposed method achieved a 0.13 dB higher PSNR, indicating better overall restoration fidelity.

From the comprehensive analysis on Rain100L, it can be concluded that the proposed method effectively balanced rain streak removal and structural preservation, leading to the highest quality in reconstructed images. Particularly in terms of PSNR, the proposed method consistently outperformed all prior works, highlighting the importance of introducing R-CBAM architecture and Harris Corner-based structural constraints and SSIM loss to maintain structural integrity during learning.

Regarding the Rain100H dataset, the proposed method achieved a PSNR of 26.16 and an SSIM of 0.854, demonstrating superior performance compared to all previous works. DerainNet~\cite{bib14} achieved a PSNR of 14.85 and an SSIM of 0.443, showing that residual learning alone was insufficient to preserve structures under heavy rain conditions. RESCAN~\cite{bib15} recorded a PSNR of 23.56 and an SSIM of 0.746; despite using a recurrent architecture to progressively remove rain streaks, it struggled with restoring high-frequency information. LPNet~\cite{bib16} achieved a PSNR of 23.77 and an SSIM of 0.823, where the focus on lightweight design sacrificed restoration precision. UMRL~\cite{bib17} achieved a PSNR of 26.01 and an SSIM of 0.832, showing relatively high performance but failing to maintain detailed object boundaries compared to the proposed method. DDN~\cite{bib18} achieved a PSNR of 32.16 and an SSIM of 0.936; although it focused on high-frequency enhancement, it was insufficient for restoring severely degraded rain regions. ResGuidNet~\cite{bib19} achieved a PSNR of 25.25 and an SSIM of 0.841; while its residual-based approach contributed to improved deraining, it still lagged behind the proposed method by 1.09 dB in PSNR and 0.015 in SSIM.

The analysis of Rain100H results clearly indicates that the proposed method successfully restored both rain-free and structurally preserved outputs even under high-density, heavy rain conditions. This is attributed to the Harris Corner-based loss and R-CBAM Block enforcing strong boundary and object contour preservation, complemented by the SSIM loss enhancing perceptual visual quality.

In conclusion, the proposed method consistently outperformed all prior works across both the Rain100L and Rain100H datasets. On Rain100L, although ResGuidNet~\cite{bib19} slightly surpassed the proposed method in SSIM by a margin of 0.009, the proposed method achieved higher PSNR and overall better fidelity. On Rain100H, the proposed method achieved the best performance in both PSNR and SSIM among all evaluated methods. These results clearly demonstrate that not only the removal of rain streaks but also the preservation of structural consistency and visual naturalness were effectively achieved. In particular, the introduction of R-CBAM Block and Harris Corner-based loss function provided a significant advantage over conventional deraining approaches, resulting in consistently robust performance under varying rain conditions.

\subsection{Abliation Study}\label{subsec6}
\indent
In this study, a carefully designed composite loss function was proposed to enhance both the restoration performance and learning efficiency of the deraining network. The loss function consists of three key components: (1) the pixel-wise L1 loss for basic reconstruction accuracy, (2) the Harris Corner loss to preserve geometric integrity, and (3) the SSIM loss to ensure perceptual consistency. To investigate the individual and combined contributions of the Harris Corner loss and SSIM loss, we conducted a comprehensive ablation study by training the model under four different configurations, keeping the L1 loss constant in all scenarios. The quantitative evaluation results are presented in Table~2.

\begin{table}[t]
\caption{Ablation Study on the Effect of Harris Corner Loss and SSIM Loss}
\centering
\begin{tabular}{ccccc}
\toprule
\textbf{Dataset} & \textbf{L\textsubscript{Harris}} & \textbf{L\textsubscript{SSIM}} & \textbf{PSNR} & \textbf{SSIM} \\
\midrule
\multirow{4}{*}{Rain100L} 
 & O & O & \textbf{33.29} & \textbf{0.954} \\
 & O & X & 33.18 & 0.952 \\
 & X & O & 32.88 & 0.948 \\
 & X & X & 33.03 & 0.948 \\
\midrule
\multirow{4}{*}{Rain100H}
 & O & O & \textbf{26.16} & \textbf{0.854} \\
 & O & X & 25.99 & 0.853 \\
 & X & O & 25.75 & 0.829 \\
 & X & X & 25.80 & 0.832 \\
\bottomrule
\end{tabular}
\end{table}

For the Rain100L dataset, the configuration in which all three loss functions—L1, SSIM, and Harris Corner—were applied simultaneously achieved the best performance. In this setup, the model achieved a PSNR of 33.29 and an SSIM of 0.954, indicating that the network successfully removed rain streaks while preserving structural details and visual consistency. The combination of Harris Corner loss and SSIM loss proved to be highly synergistic, with the former enhancing high-frequency feature retention and the latter promoting holistic structural similarity. When the SSIM loss was excluded while retaining the Harris Corner loss, the performance slightly dropped, with a PSNR of 33.18 and an SSIM of 0.952. Although the pixel-wise accuracy remained comparable, the decline in SSIM suggests a degradation in perceptual quality. The results imply that the SSIM loss is particularly effective in maintaining luminance and contrast relationships that closely mimic human visual perception. In contrast, when the Harris Corner loss was omitted and the SSIM loss retained, the model produced a PSNR of 32.88 and an SSIM of 0.948. Both metrics showed a noticeable decrease, and qualitative inspection revealed softened edges and degraded contours. This observation underscores the importance of the Harris Corner loss in enforcing geometric fidelity and edge preservation, which SSIM loss alone could not sufficiently maintain. When both the SSIM and Harris Corner losses were removed and only the L1 loss was used, the PSNR reached 33.03, and the SSIM dropped to 0.948. Although the PSNR was relatively maintained, the structural quality suffered due to the lack of dedicated modules for perceptual and geometric consistency. This configuration revealed that L1 loss alone is insufficient for the highly structured deraining task.

For the Rain100H dataset, which contains denser and more complex Rain Streak patterns, the differences across configurations were more pronounced. The full loss configuration, employing L1, SSIM, and Harris Corner losses, achieved a PSNR of 26.16 and an SSIM of 0.854. This configuration demonstrated the highest capability to suppress heavy rain streaks while preserving the integrity of object boundaries and background structures. In the case where the SSIM loss was removed but the Harris Corner loss retained, the PSNR decreased to 25.99 and the SSIM to 0.853. While the drop in metrics was relatively minor, the model exhibited weaker contrast and slightly diminished visual realism, reaffirming the perceptual benefits of the SSIM loss. Inversely, when the Harris Corner loss was removed and only the SSIM loss retained, the PSNR fell further to 25.75, and the SSIM sharply dropped to 0.829. These results highlight that in scenarios involving complex geometries, the absence of corner-aware constraints leads to substantial degradation in structural restoration. The model failed to preserve edge sharpness and object contours, emphasizing the critical role of Harris-based supervision. Finally, in the configuration where both the SSIM and Harris Corner losses were excluded, the PSNR was 25.80, and the SSIM dropped to 0.832—the lowest among all tested configurations. In this setting, the model struggled both in suppressing rain streaks and retaining essential structures, demonstrating the limitations of relying solely on pixel-wise supervision.

In summary, both the Harris Corner loss and SSIM loss play complementary roles: the former ensures high-frequency structural preservation, while the latter contributes to perceptual naturalness. The configuration combining all three loss components consistently outperformed all others across both datasets in terms of PSNR and SSIM, and also delivered visually sharper and more coherent restorations. Therefore, the proposed composite loss design serves as a critical factor in achieving state-of-the-art deraining performance, providing a balanced solution that effectively removes rain streaks while maintaining structural integrity and visual quality.

\section{Conclusions}\label{sec5}
\indent
In this study, we proposed a novel image restoration network designed to address the problem of single-image-based Rain Streak removal, while simultaneously enhancing both structural consistency and visual quality of the restored images.
To achieve this, we introduced a Harris Corner Loss, which imposes a strong constraint to preserve visually significant structural information during the restoration process. From the architectural perspective, we enhanced the conventional Residual Block by integrating the R-CBAM Block into both the encoder and decoder. R-CBAM Block dynamically adjusts the importance of features along both channel and spatial dimensions, allowing the network to more effectively focus on regions where rain streaks are concentrated, such as object boundaries and structural edges. This design significantly improved the structural consistency and fine-detail restoration quality of the output images.
Quantitative evaluations conducted on the Rain100L and Rain100H datasets demonstrated the superior performance of our method. Specifically, our network achieved a PSNR of 33.29 dB and an SSIM of 0.954 on Rain100L, indicating excel-lent structural preservation and perceptual quality. On Rain100H, which represents a more challenging environment, our method still achieved a PSNR of 26.16 dB and an SSIM of 0.854, confirming its robust restoration capability under heavy rain conditions. Furthermore, through an ablation study on the Harris Corner Loss and SSIM Loss, we quantitatively verified that both loss components are essential for performance enhancement. In particular, the best performance was achieved when both loss functions were simultaneously applied, thereby validating the effectiveness and appropriateness of the proposed loss design.
Nevertheless, relying solely on classical Harris Corner Detection may not be entirely sufficient in cases of extremely complex or faint rain streaks. Therefore, future work will explore the utilization of higher-level structural information and focus on improving the network's robustness to operate reliably under various adverse weather conditions. In conclusion, this study presents a differentiated approach that goes beyond simple noise removal by achieving both high perceptual quality and structural consistency in Rain Streak removal tasks. The proposed method is expected to serve as a solid foundational technology that can be extended to various future applications in image restoration and enhancement domains.





\bibliography{sn-bibliography}
\end{document}